%% file: root.tex
\documentclass[letterpaper, 10 pt, conference]{ieeeconf}  % Comment this line out if you need a4paper

\IEEEoverridecommandlockouts                              % This command is only needed if 
\usepackage{graphicx}%
\usepackage{multirow}%
\usepackage{array}
\usepackage{amsmath,amssymb,amsfonts}%
\usepackage{mathrsfs}%
\usepackage{dblfloatfix}
\usepackage{cite}
\usepackage{caption}
\usepackage{subcaption}
\usepackage{color,soul}
\usepackage{pdfpages}

\title{\LARGE \bf
Adaptive Control for Triadic Human-Robot-FES Collaboration in Gait Rehabilitation: A Pilot Study
}

\author{Andreas Christou$^{1}$, Antonio J. del-Ama$^{2}$, Juan C. Moreno$^{3}$ and Sethu Vijayakumar$^{4}$% <-this % stops a space
\thanks{*This research was supported by the Engineering and Physical Sciences Research Council (EPSRC, grant reference EP/L016834/1) as part of the Centre for Doctoral Training in Robotics and Autonomous Systems at Heriot-Watt University and The University of Edinburgh, by the Honda Research Institute Europe and by the Alan Turing Institute, U.K. This work is also part of the Spanish research Network “FUSION” (RED2022-134319-T) funded by MCIN/AEI /10.13039/501100011033.}% <-this % stops a space
\thanks{$^{1}$Andreas Christou is with the School of Informatics, University of Edinburgh, UK,
{\tt\small andreas.christou@ed.ac.uk}}%
\thanks{$^{2}$Antonio J. del-Ama is with the Electronic Technology Department, Rey Juan Carlos University, Spain
}%%
\thanks{$^{3}$Juan C. Moreno is with the Department of Translational Neuroscience, Cajal Institute, Spain
}%%
\thanks{$^{4}$Sethu Vijayakumar is with the School of Informatics, University of Edinburgh, UK, and with the Alan Turing Institute, UK
}%%
}

\begin{document}

\maketitle
\thispagestyle{empty}
\pagestyle{empty}

%%%%%%%%%%%%%%%%%%%%%%%%%%%%%%%%%%%%%%%%%%%%%%%%%%%%%%%%%%%%%%%%%%%%%%%%%%%%%%%%
\begin{abstract}
The hybridisation of robot-assisted gait training and functional electrical stimulation (FES) can provide numerous physiological benefits to neurological patients. However, the design of an effective hybrid controller poses significant challenges. In this over-actuated system, it is extremely difficult to find the right balance between robotic assistance and FES that will provide personalised assistance, prevent muscle fatigue and encourage the patient's active participation in order to accelerate recovery. In this paper, we present an adaptive hybrid robot-FES controller to do this and enable the triadic collaboration between the patient, the robot and FES. A patient-driven controller is designed where the voluntary movement of the patient is prioritised and assistance is provided using FES and the robot in a hierarchical order depending on the patient's performance and their muscles' fitness. The performance of this hybrid adaptive controller is tested in simulation and on one healthy subject. Our results indicate an increase in tracking performance with lower overall assistance, and less muscle fatigue when the hybrid adaptive controller is used, compared to its non adaptive equivalent. This suggests that our hybrid adaptive controller may be able to adapt to the behaviour of the user to provide assistance as needed and prevent the early termination of physical therapy due to muscle fatigue.

\end{abstract}

%%%%%%%%%%%%%%%%%%%%%%%%%%%%%%%%%%%%%%%%%%%%%%%%%%%%%%%%%%%%%%%%%%%%%%%%%%%%%%%%
\section{INTRODUCTION}
In the last two decades, more and more research has focused on deploying technological advancements in physical therapy in order to accelerate learning and reduce the recovery time of people with neurological disorders. To alleviate the strain on physical therapists, the use of robotic orthoses and wearable exoskeletons has been studied \cite{Campagnini2022,Shi2019a}. These devices have the potential to provide body weight support and quantitatively assess the performance of the patient in order to provide systematic assistance. In some cases, these devices were found to be beneficial in improving gait speed and balance in people with spinal cord injury \cite{Fang2020,Hayes2018} as well as independent gait in stroke survivors \cite{Tedla2019,Barbuto2019a,Swinnen2014}. 

\begin{figure*}[ht]
    \centering
    \includegraphics[width=0.85\textwidth]{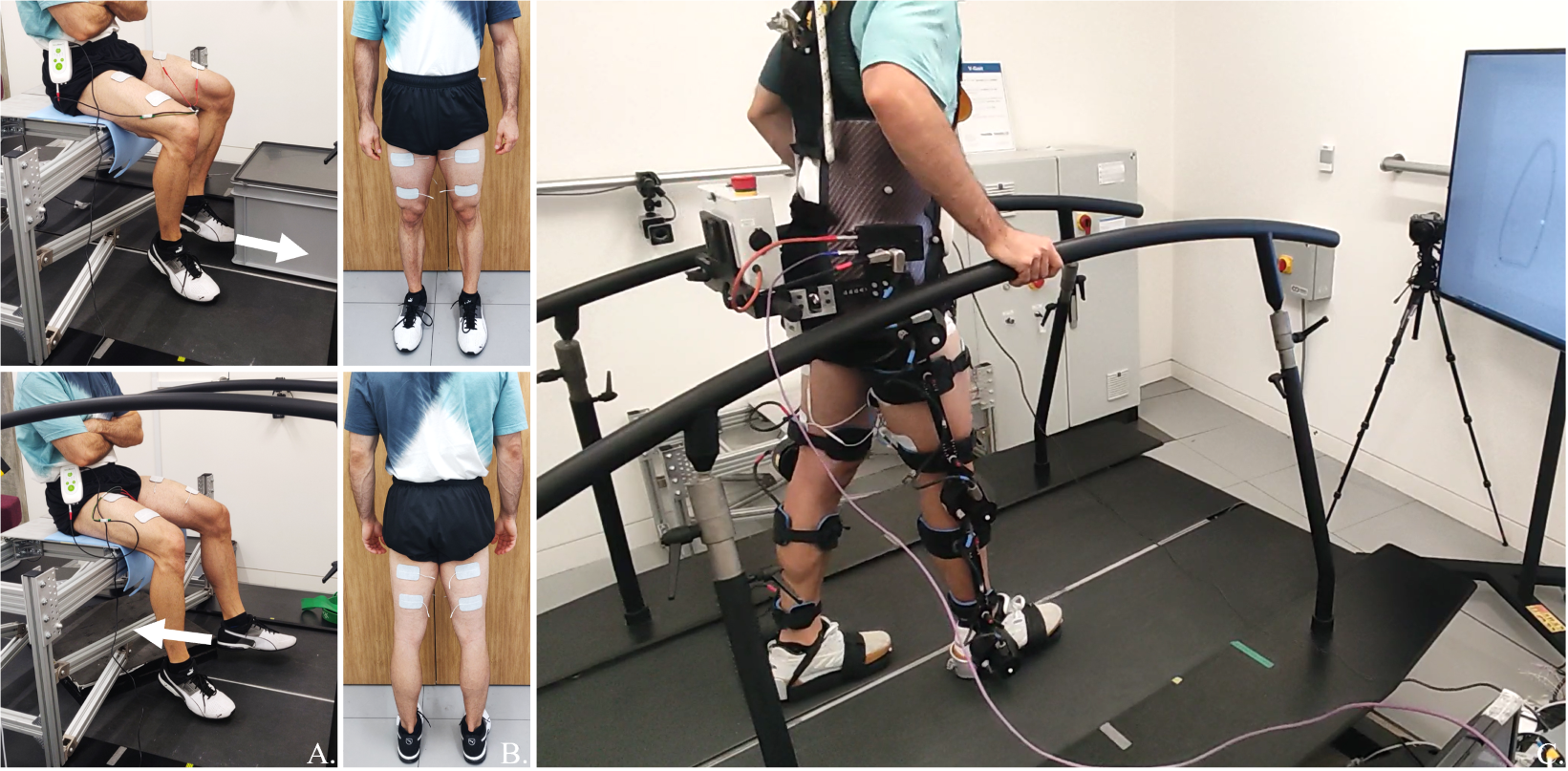}
    \caption{A. Fatigue model identification for quadriceps and hamstrings using FES in isometric conditions (pushing a heavy box on an instrumented treadmill). B. Electrode placement. C. Experimental validation of hybrid adaptive controller.}
    \label{fig:triadic_collaboration}
\end{figure*}

To further enhance the outcomes of physiotherapy, the combination of robotic assistance with functional electrical stimulation (FES) has been proposed \cite{Anaya2018,Del-Ama2012}. The use of FES can induce muscle contractions and utilise the patient's muscles as actuators. This can provide increased calcium concentration in bones and improved cardiovascular benefits \cite{Peckham2005}. Also, the afferent feedback to the patient is increased, which is associated with increased neuroplasticity \cite{Marquez-Chin2020,PopovicM2016}. However, with the use of FES the rapid onset of muscle fatigue \cite{Ibitoye2016b,Bickel2011} and the non-linear response of muscles make the induced motion hard to control \cite{Lynch2008,Schauer2017}. When combined with robotic assistance (Fig. \ref{fig:triadic_collaboration}), these limitations can be overcome but achieving a seamless triadic collaboration between the user, the robot and FES remains a challenge.

Some of the earlier studies on hybrid systems tested the possibility of combining FES with passive wearable orthoses \cite{Goldfarb1996, Kobetic2009,Sharma2014}. However, the passive nature and the dissipative control of these orthoses, implied that the joint torque induced by FES needed to match or exceed the desired joint torque. Thus, even though these passive orthoses proved to be safe, their inability to be actively controlled to provide assistive torques that would increase the energy of the system limited their capacity to reduce muscle fatigue.

To address this, more emphasis was given on combining FES with active electromechanical devices. This led to the exploration of a more diverse pool of assistive controllers with the majority of them incorporating adaptive control for either or both the robot and FES. Examples include the studies by Stauffer et al. \cite{Stauffer2009b} and Ha et al. \cite{Ha2016d} where iterative learning control (ILC) was used to adjust the stimulation intensity of FES in order to minimise the interaction torques between the user and the robot while performing an ambulatory task. ILC was also used in \cite{Del-Ama2014a} for both the FES controller and the robotic controller. Del-Ama et al. used a dual state controller that switched between a learning state and a monitoring state to enable the adjustment of the stimulation parameters and the stiffness of the exoskeleton, respectively, in order to respond to muscle fatigue, reduce the interaction between the human and the robot, and maintain an accurate tracking of the desired kinematic trajectory \cite{Del-Ama2014a}. 

Studies on hybrid robot-FES cooperation have also been carried out using optimal control and nonlinear control to address the actuation redundancy problem \cite{Kirsch2017,Molazadeh2021,Alibeji2018a}. In \cite{Kirsch2017} the use of non-linear model predictive control (NMPC) was used during a knee extension task and demonstrated good tracking performance and disturbance rejection. Similarly, in \cite{Molazadeh2021} MPC was used to distribute the assistive forces between a robotic device and FES. These assistive forces were estimated using a neural network-based ILC (NNILC) which was designed to learn the unknown system dynamics. This system was shown to be effective in reducing the tracking error by reducing the feedback input and increasing the contribution of the learning terms. A slightly different approach was adopted by Alibeji et al. \cite{Alibeji2018a} where postural synergies were used to reduce the number of control variables. Based on the identified postural synergies, the estimated muscle fatigue and a Lyapunov-based stability approach, a non-linear controller was derived which achieved accurate trajectory tracking. However, the controller's effect on muscle fatigue could not be analysed due to the limited duration of the experiments. Some interesting torque distribution methods have also been explored in simulation in order to address this actuation redundancy in hybrid control but their effectiveness was not validated experimentally \cite{Vallery2005,Romero-Sanchez2019a}. 

A common element among these controllers is that the voluntary contributions of the human are often neglected or are dealt with as disturbance to the system. However, many studies have highlighted the importance of the patient's engagement in neural plasticity and neurorehabilitation \cite{Blank2014, Paolucci2012, Kahn2006,Kaelin-Lane2005}. Another common element among these studies is that muscle fatigue is often treated as a discrete event instead of a continuous process, the management of which is usually compensatory and not preventive. As a result, increased stimulation intensity is often used to compensate for the reduced force generated by the muscles, which is likely to induce even higher levels of muscle fatigue. 

In this study, we design an adaptive hybrid controller for ambulatory patients (i.e. people who retain some voluntary control over their lower limbs) to achieve a triadic collaboration between the human, the robot and FES \cite{Gordon2023}, where the main driver of the system is the patient, and muscle fatigue is addressed in a continuous and preventive fashion. In this controller, we use a hierarchical structure to prioritise the contributions of the three agents in the following order: (1) the voluntary contributions of the patient, (2) the assistance from FES and (3) the assistance from the robot. We use a model-based approach to estimate the level of induced muscle fatigue and the controller's parameters are adapted to provide personalised assistance. This adaptation happens based on the user's estimated level of muscle fatigue and their ability to follow the desired path. The performance of this hybrid adaptive path controller (HAPC) is verified in simulation using our offline model-based approach \cite{Christou2022} and it is validated on one healthy subject. The ability of the controller to adapt to the needs of the user in order to provide assistance as needed and prevent muscle fatigue is analysed and discussed. 

\section{HYBRID ADAPTIVE PATH CONTROLLER}
\subsection{Hybrid Path Control}
\begin{figure}[h]
    \centering
    \includegraphics[width=0.99\linewidth]{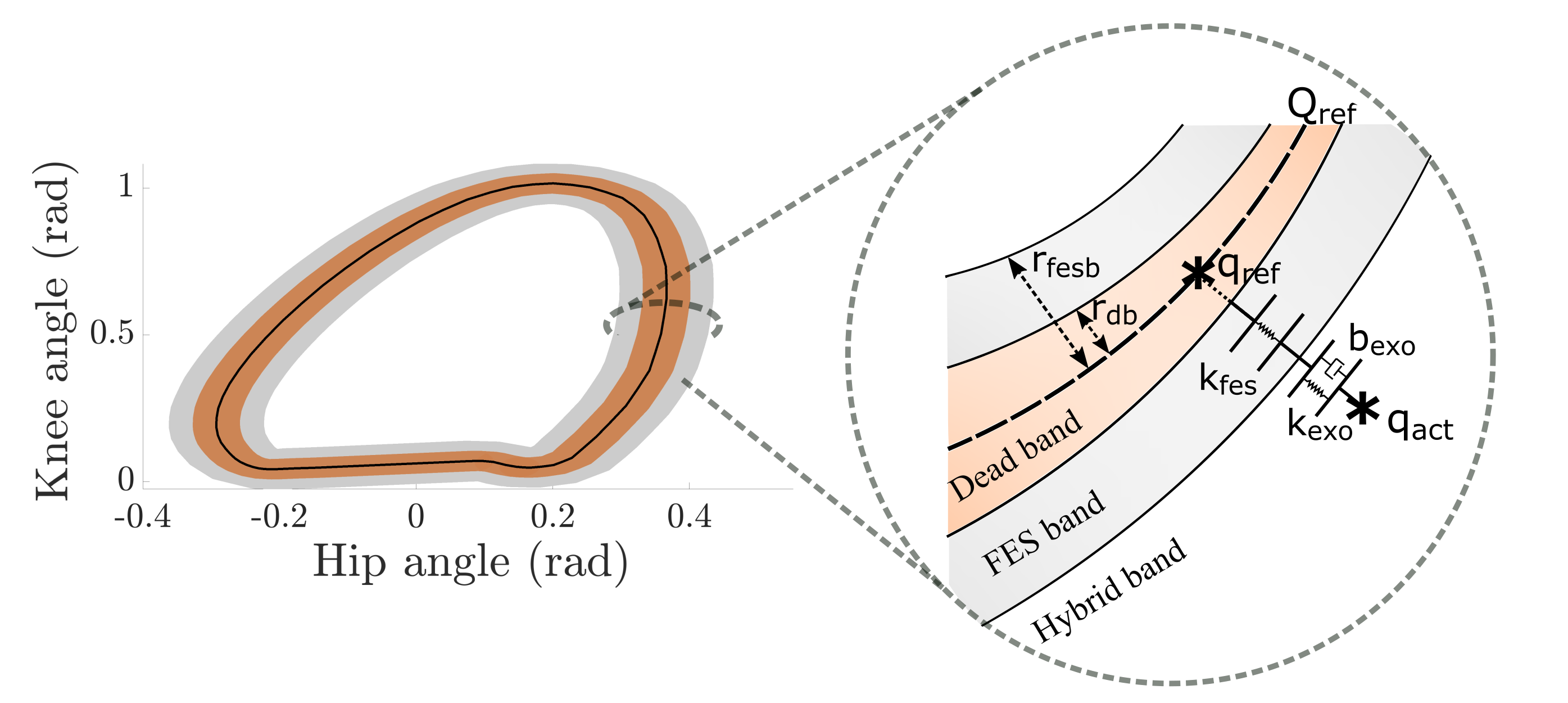}
    \caption{Illustration of the reference kinematic path in joint space, surrounded by the dead band, and the FES band. The magnified region depicts the components of the hybrid path controller for a human configuration, $q_{act}$.}
    \label{fig:hybrid_path}
\end{figure}

As our starting point we deploy a Hybrid Path Controller -- a controller that follows the principles of Path Control \cite{Duschau-Wicke2010b}, and provides assistance using a wearable robot and FES. This controller involves a reference kinematic path in joint space, ${\bf Q}_{ref} \in \mathbb{R}^{i \times j}$, that describes the desired relationship between the hip and the knee joint angle in the sagittal plane (where $i$ is the number of points in the discretised domain of the reference path and $j$ is the number of degrees of freedom defined by the reference path). Based on this reference path, the reference point, ${\bf q}_{\text{ref}} \in \mathbb{R}^{j}$, is calculated according to the joint angles of the human, ${\bf q}_{\text{act}}  \in \mathbb{R}^{j}$ (for this study, the joint angles of the human are approximated as the joint angles of the exoskeleton). Given the joint angles of the human, this reference point is dynamically defined as the point on the reference path where the Euclidean distance between the reference path, ${\bf Q}_{ref}$, and the joint angles of the human, ${\bf q}_{act}$, is at a minimum (Fig. \ref{fig:hybrid_path}). To establish a hierarchy between the human, the FES and the robot, a dead band and a FES band of radius, $r_{db}$ and $r_{fesb}$, respectively, are defined around the reference path. This defines a region where no assistance is provided to the human (dead band), a region where only assistance from FES is provided (FES band) and a region where hybrid robot-FES assistance is provided (hybrid band). The corresponding joint angle error can then be calculated for the FES controller, ${\Delta {\bf q}}_{f} \in \mathbb{R}^{j}$, and the exoskeleton's controller, ${\Delta {\bf q}}_{e} \in \mathbb{R}^{j}$, as the difference between the reference point and the human's pose, minus the relevant error tolerance, $r$, formally expressed as:
\begin{gather}
    {\boldsymbol \epsilon}= {{\bf q}}_{\text {ref}} - {\bf q}_{\text {act}}, \label{eq:pathControl1}\\
    {\Delta {q}}^{(j)} =
        \begin{cases}
            0 , & |{\epsilon}^{(j)}| \leq r, \\
            {\epsilon}^{(j)} - r , & {\epsilon}^{(j)} > r,  \\
            {\epsilon}^{(j)} + r, & {\epsilon}^{(j)} < -r,
        \end{cases} \\
    r=
    \begin{cases}
        r_{\text{db}}, \hspace{10pt} \text{FES controller},\\
        r_{\text{fesb}}, \hspace{10pt} \text{Exo controller}.\\
    \end{cases}
\end{gather} 
Based on the tracking error, assistance is provided by the FES and the robot, in a direction orthogonal to the reference path in order to promote patient-driven therapy without temporal constraints. For this study the reference path was defined based on the recorded motion of a healthy subject and was adjusted to a path with a less pronounced loading response.

\subsection{Adaptive FES Controller} \label{sec:AdaptiveFes}
To enable the hybrid controller to adjust the intensity of electrical stimulation to the needs of the user, a proportional controller with two adaptive components is defined as:
\begin{gather}
    {\bf u}_{f} = {\boldsymbol \mu}{\boldsymbol \gamma}{\bf K}_{f}\Delta{\bf \overline q}_{f}, \label{eq:FES_adaptive_control}\\
    {\bf K}_{f}=\frac{({{\bf u}_{sat}-{\bf u}_{thr}})}{2{r}_{\text{fesb}}^{0}},
\end{gather}
where ${\bf{u}}_{f} \in \mathbb{R}^{k}$ is the stimulation intensity, $\boldsymbol{\mu} \in \mathbb{R}^{k}$ is a measure of muscle fitness (the inverse of muscle fatigue) ranging from [0,1], $\boldsymbol{\gamma} \in \mathbb{R}^{k}$ is the ILC gain ranging from [0,1], ${\bf{K}}_{f} \in \mathbb{R}^{k}$ is a constant stiffness, and ${\Delta \overline {\bf q}}_{f} \in \mathbb{R}^{k}$ is the kinematic error adjusted by its sign and the stimulated muscles to ensure ${\bf u}_{f}\geq 0$, with $k$ being the number of stimulated muscles. ${\bf u}_{thr}$ and ${\bf u}_{sat}$ are the threshold and saturation pulse width for the given stimulation amplitude and frequency, and $r_{\text{fesb}}^{0}$ is the initial radius of the FES band (set to 6\textdegree). 

The first adaptive component involves the adjustment of the stimulation intensity according to muscle fatigue. That is, muscles whose fatigue increases receive less stimulation. To estimate muscle fatigue the following model is used \cite{Riener1998}:
\begin{gather}
    {\dot \mu} = \frac{({\mu}_{min} - \mu)\rho(f)a_{f}}{T_{fat}} + \frac{(1-\mu)(1-\rho(f)a_{f})}{T_{rec}},  \label{equation:RienerFitnessCurve}\\
    \rho(f) = 1-\beta + \beta{(\frac{f}{100})}^{2}, \ \text{for} \ f<100 \ Hz,
\end{gather}
where $\mu_{min}$ is the minimum muscle fitness, $\beta$ is a shape factor, $f$ is the stimulation frequency, and $T_{fat}$ and $T_{rec}$ are the time constants that describe the rate of muscle fatigue and recovery, respectively. The parameter $a_{f}$ is the activation of the stimulated muscles which is estimated as \cite{GFOHLER2004, Romero-Sanchez2019a}: 
\begin{gather}
    {\bf e}=
    \begin{cases}
        0, & {\bf u}_{f}<{\bf u}_{thr}, \\
        \frac{{\bf u}_{f}-{\bf u}_{thr}}{{\bf u}_{sat}-{\bf u}_{thr}}, & {\bf u}_{thr}<{\bf u}_{f}<{\bf u}_{sat},\\
        1, & {\bf u}_{f}>{\bf u}_{sat}. 
    \end{cases} \\
    {\bf k}_{1}{\bf \ddot{a}} + {\bf k}_{2}{\bf \dot{a}} + {\bf a} = {\bf e}, \\
    {\bf a}_{f}={\bf a}{\boldsymbol{\mu}}
\end{gather}
where ${\bf k}_{1}$ and ${\bf k}_{2}$ are constants that characterise the response of muscles to electrical stimulation as described in \cite{GFOHLER2004} including a rise time constant, $T_{rise}$, a fall time constant, $T_{fall}$, and the excitation time constant, $T_{e}$, where $k_1=T_{e}*T$, and $k_2=T_{e}+T$ for $T=T_{rise}$, if $e>a$ and $T=T_{fall}$, otherwise. 
\begin{table*}[ht]
\caption{Identified parameters for muscle fatigue model}
\label{table:FatigueModel}
\centering
\begin{tabular}{ |p{2mm}|c|c|c|c|c|c|c|c|c|c| } 
\hline
 & & $u_{thr}(\mu s)$ & $u_{sat}(\mu s)$ & $T_{fat}(s)$ & $T_{rec}(s)$ & $T_{rise}(s)$ & $T_{fall}(s)$ & $T_{e}(s)$ & $\mu_{min}$ & $\beta$ \\
\hline
\parbox[t]{2mm}{\multirow{2}{*}{\rotatebox[origin=c]{90}{Right}}}& Quadriceps & 100 & 700 & 57.01 & 59.87 & 0.2071 & 0.1370 & 0 & 0.07 & 0.0747 \\
\cline{3-11}
 & Hamstring & 250 & 600 & 64.34 & 65.27 & 0.2440 & 0.0829 & 0.06 & 0.13 & 0.1493\\
 \hline
 \parbox[t]{2mm}{\multirow{2}{*}{\rotatebox[origin=c]{90}{Left}}}& Quadriceps & 200 & 600 & 36.05 & 69.56 & 0.1428 & 0.2533 & 0 & 0.17 & 0.2453\\
 \cline{3-11}
 & Hamstring & 250 & 550 & 44.58 & 105.19 & 0.1963 & 0.1797 & 0.002 & 0.14 & 0.2347 \\
 \hline
\end{tabular}
\end{table*}

The second adaptive component of this controller is the stimulation adjustment that occurs iteratively, at every gait cycle, based on the root mean squared error of the user. This adaptive component reduces the value of parameter $\gamma$ as the performance of the patient improves until the patient can perform the task without any assistance. For this, a finite state machine (FSM) is used to differentiate between the swing phase (SW) and stance phase (ST) where different assistive forces may be appropriate. This is defined as:
\begin{gather}
    {\boldsymbol \gamma}^{z+1}_{st} = {\boldsymbol \phi}_{f}{\boldsymbol \gamma}^{z}_{st} + {\boldsymbol \lambda}_{f}{\text{rms} ({\Delta \bf \tilde q}_{f})}^{z}_{st},\\
    {\boldsymbol \gamma}^{z+1}_{sw} = {\boldsymbol \phi}_{f}{\boldsymbol \gamma}^{z}_{sw} + {\boldsymbol \lambda}_{f}{\text{rms} ({\Delta \bf \tilde q}_{f})}^{z}_{sw}, \\
    \lambda_{f}=(1-\phi_{f}),
\end{gather}
where ${\boldsymbol{\phi}_{f}}$ and $\boldsymbol{\lambda}_{f}$ are the forgetting and learning factors of the FES ILC (${\boldsymbol \phi}_{f}$ set to 0.95), ${\Delta \bf \tilde q}_{f}$ is the error normalised by $r_{db}$ (set to 2\textdegree) and $z$ is the number of gait cycles performed.

\subsection{Adaptive Exoskeleton Controller} \label{sec:AdaptiveExo}
For the robotic assistance, an adaptive proportional-derivative controller is used. This is defined as: 
\begin{gather}
    {\boldsymbol \tau}_{e} = {\bf K}_{e}{\Delta{{\bf q}}}_{e} + {\bf B}{\Delta{{\dot{\bf q}}}}_{e}, \label{eq:impedanceController1} \\
    {\bf B}={\bf c}_{cr}\sqrt{\bf K}, \label{eq:impedanceController2}
\end{gather}
where ${\bf K}_{e}$ and ${\bf B}_{e}$ are the stiffness and damping of the exoskeleton's joints, respectively, and ${\bf c}_{cr}$ is the matrix of the critical damping coefficients. 

Like the FES controller, the exoskeleton controller also uses a FSM in order to adjust the stiffness of the exoskeleton based on the phase in the gait cycle. On every gait cycle, the root mean squared error of the user is recorded and is used to update the exoskeleton's stiffness as follows: 
\begin{gather}
    {\bf K}^{z+1}_{st}= {\boldsymbol \phi}_{e}{\bf K}^{z}_{st}+ {\boldsymbol \lambda}_{e}{\bf K}_{e}^{0}{\text{rms}({\Delta \bf \tilde q}_{e})}^{z}_{st}, \\
    {\bf K}^{z+1}_{sw}= {\boldsymbol \phi}_{e}{\bf K}^{z}_{sw}+{\boldsymbol \lambda}_{e}{\bf K}_{e}^{0}{\text{rms}({\Delta \bf \tilde q}_{e})}^{z}_{sw}, \\
    \lambda_{e}= (1-\phi_{e}),
\end{gather}
where ${\boldsymbol \phi}_{e}$ and ${\boldsymbol \lambda}_{e}$ are the forgetting and the learning factors of the exoskeleton ILC (${\boldsymbol \phi}_{e}$ set to 0.95), and ${\bf K}_{e}^{0}$ is the baseline stiffness (set to 340 Nm/rad \cite{Del-Ama2014a}). 

\subsection{Adaptive FES Band} \label{sec:AdaptiveTunnel}
The adaptation of the width of the FES band is designed in order to achieve the collaboration between the robot and the FES. By reducing the width of the FES band according to the fitness level of the muscles, robotic assistance is combined with FES at an earlier stage. This adaptation of the FES band is implemented as:
\begin{gather}
    {r}_{\text{fesb}} = {r}_{\text{fesb}}^{{0}}{\tilde{\mu}},
\end{gather}
where $\tilde{\mu}$ is the mean fatigue across the stimulated muscles.

\section{CONTROLLER VALIDATION}
\subsection{Subject}
The effectiveness of the HAPC was tested both in simulation and experimentally on one healthy volunteer (male, age = 30, weight = 80 kg). 
The experiment pipeline was approved by the University of Edinburgh, School of Informatics Ethics Committee (ID 2021/46920) and the participant provided written consent.
\begin{figure*}[h]
    \centering
    \includegraphics[width=0.9\textwidth]{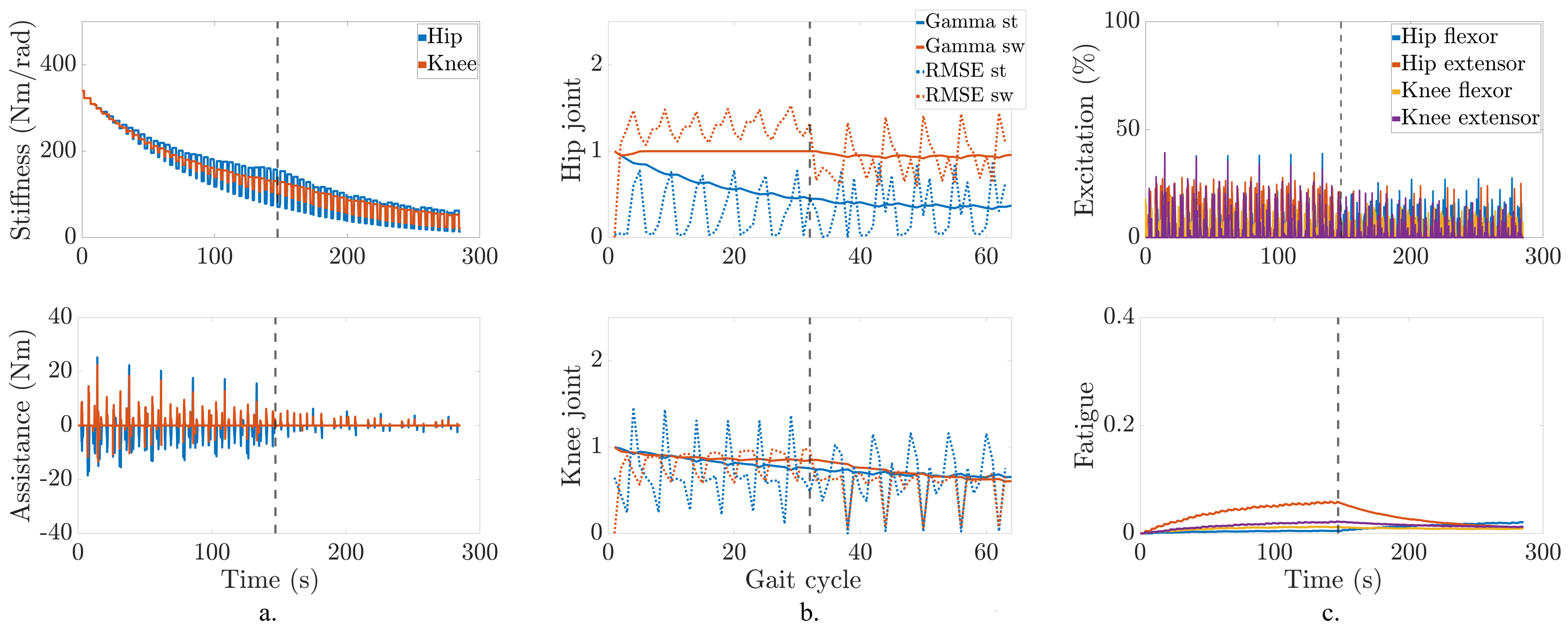}
    \caption{Simulated response of the hybrid controller. (a) Shows the robotic stiffness and assistance provided, (b) the trajectory error and adaptation of parameter $\gamma$, and (c) the muscle fatigue and stimulation provided with the adaptive controller during poor model performance (before grey dotted line) and more accurate performance (after grey dotted line).}
    \label{fig:sim_results}
\end{figure*}
\begin{figure}[h]
    \centering
    \includegraphics[width=0.85\linewidth]{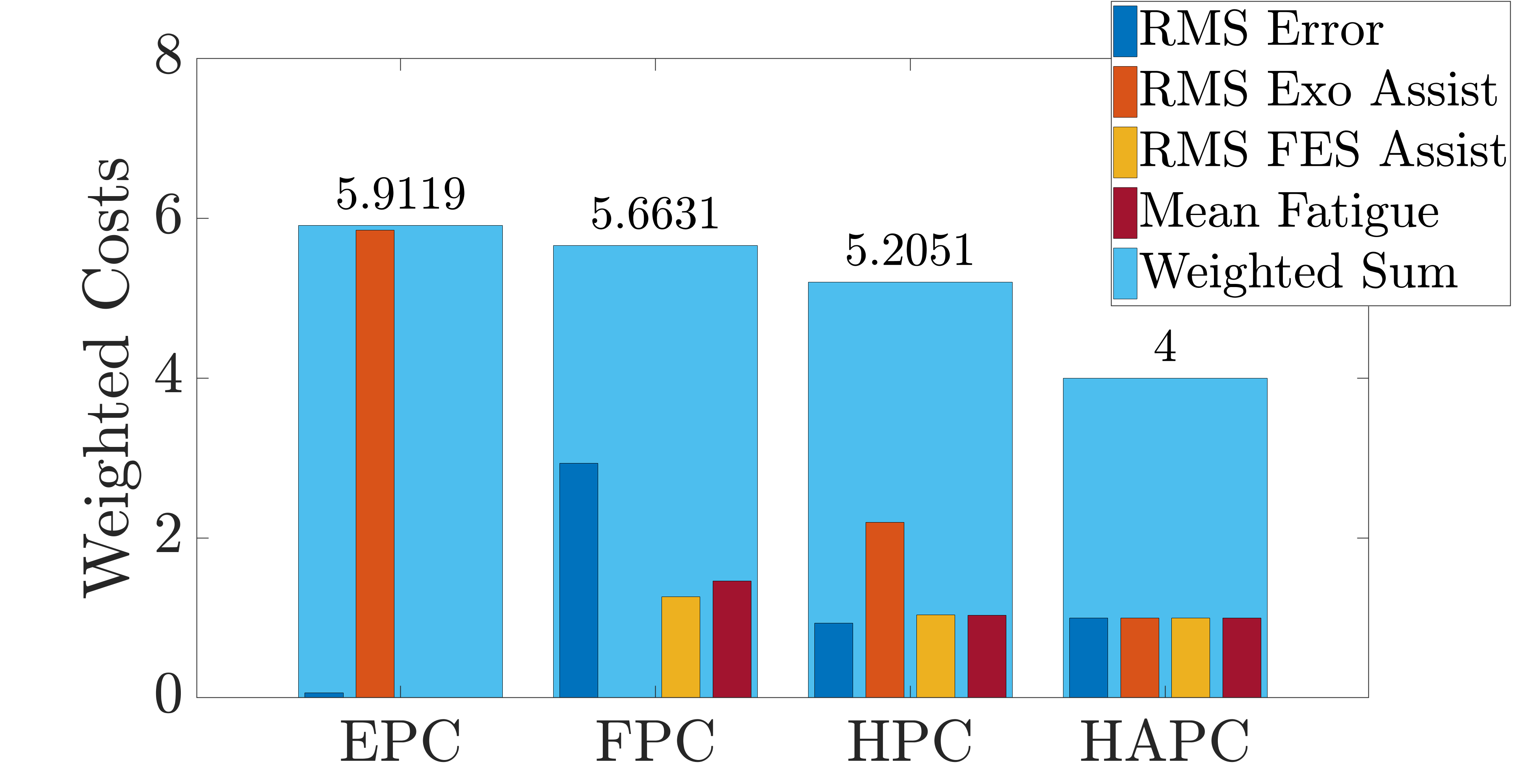}
    \caption{Comparison of the normalised trajectory error, robotic assistance, FES, and muscle fatigue between the exoskeleton controller (EPC), the FES controller (FPC), the hybrid controller (HPC) and the hybrid adaptive controller (HAPC).}
    \label{fig:sim_results_adaptation_comparison}
\end{figure}

\subsection{Hardware}
The RehaMove3 stimulator (Hasomed, Germany) was used to provide FES using rectangular surface electrodes (5x9 cm). The instrumented treadmill M-Gait (Motek Medical, Netherlands) was used to facilitate the fatigue model identification (section \ref{sec:FatigueModelCalibration}) and to enable self-paced gait during the experiment. The exoskeleton Exo-H3 (Technaid, Spain) was used to provide assistance during the task and the exoskeleton's joint position sensors were used to record the joint angle of the legs and provide visual feedback to the user (Fig. \ref{fig:triadic_collaboration}c). Simulink Desktop Real Time was used for the real-time control of the exoskeleton and FES at 100 Hz.  

\subsection{Fatigue Model Identification} \label{sec:FatigueModelCalibration}
To ensure that the estimated fatigue of muscles reflects the fitness level of the participant, a model calibration was carried out as described in \cite{Riener1996,Sharma2017b}. The feet of the participant were placed in contact with a heavy (immovable) box and the generated muscle force (contact force) was measured using the treadmill's force sensors. A symmetric biphasic signal with constant stimulation frequency of 25 Hz and current amplitude of 25 mA was used under isometric conditions (Fig. \ref{fig:triadic_collaboration}a). Firstly, to identify the minimum stimulation pulse width that results in a visible muscle contraction, $u_{thr}$, and the pulse width above which the generated force saturates, $u_{sat}$, a series of signals were delivered at 50 $\mu s$ increments until a pulse width of 800 $\mu s$ was reached. The signal duration was set to 4 s, with a resting period of 20 s. A fatigue test was then carried out using a continuous signal at the saturation intensity for 180 s, followed by a 2-minute recovery test which included pulses of 1 s, with a between-pulse rest of 10 s. The identification of the model parameters was carried out using MATLAB's optimisation tool \textit{fmincon}. The identified parameters are presented in Table \ref{table:FatigueModel}.

\begin{figure*}[h]
    \centering    
    \includegraphics[width=0.9\linewidth]{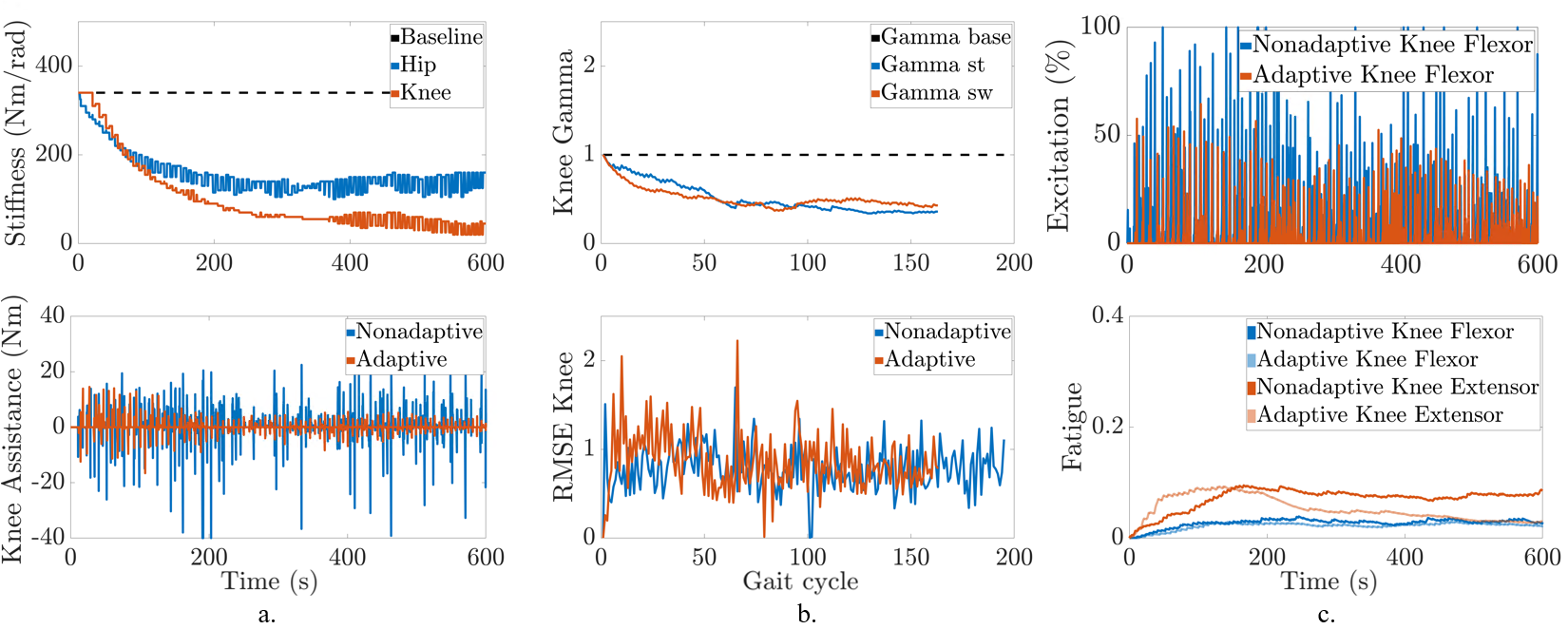}
    \caption{Experimental response of hybrid controller. Comparison between (a) the robotic stiffness and assistance, (b) trajectory error and (c) muscle fatigue and stimulation provided with the adaptive and non adaptive controllers for the left leg.}
    \label{fig:HPC_ExpResults}
\end{figure*}
\begin{figure}[h]
    \centering \includegraphics[width=0.833\linewidth]{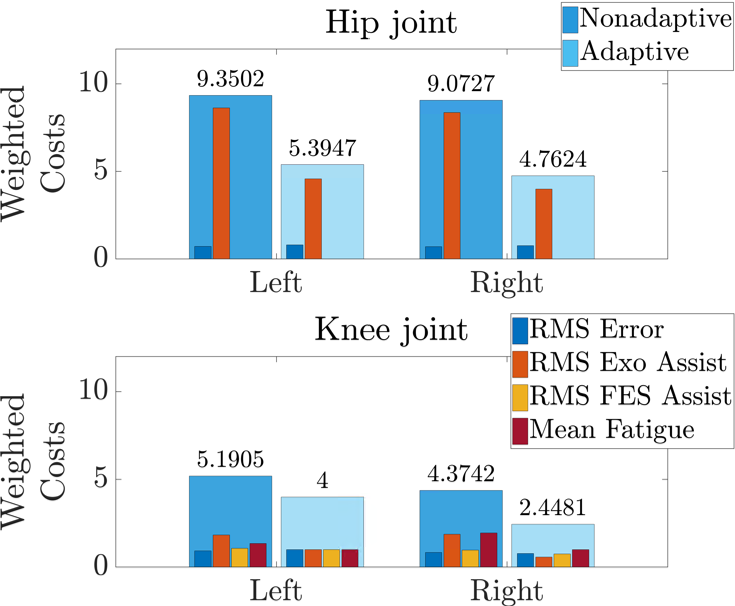}
    \caption{Comparison between the non adaptive and the adaptive controller for the recorded data of the hip and knee joints for both legs (normalised by the data recorded from the left knee joint during the adaptive control).}
    \label{fig:S2_WeightedCostsResultsExperiment}
\end{figure}

\subsection{Simulation Setup} \label{sec:sim_and_modelling}
The initial development and testing of the controller was carried out using musculoskeletal modelling. Using the biomechanical modelling software OpenSim \cite{Seth2018}, a generic lower-limb model, \textit{gait1018}, was scaled to the dimensions of the participant and was combined with the H3 exoskeleton as described in \cite{Christou2022}. An estimate of the voluntary joint torques needed for the tracking of the trajectory was obtained using motion capture and forward dynamics simulations were carried out to predict the tracking error, muscle fatigue and assistance provided to the model \cite{Christou2022}. The exoskeleton assistance was simulated using ideal joint actuators while an estimate of the FES-induced torques was obtained based on Hill-type muscle models \cite{Millard2013} and the activation dynamics described in section \ref{sec:AdaptiveFes} for the monoarticular muscles of both the hip joint and the knee joint. 

To observe the controller's response to a changing human behaviour, simulations were carried out with two different inputs for the human controls. For the first half of the simulation, a behaviour that corresponds to a high-error trajectory was used, while for the second half, a behaviour that corresponds to a low-error trajectory was used. These trajectories were obtained from our previous study in \cite{Christou2022}. Using these model controls, a comparison was also carried out between the hybrid adaptive path controller (HAPC), its non adaptive equivalent (HPC), and the exoskeleton-only (EPC) and FES-only controllers (FPC). In all cases, 64 cycles were simulated, for which the RMS error, RMS of the assistance and mean fatigue were used for comparison.

\subsection{Experimental Setup}
During experimental validation of the controller, the participant was fitted with the exoskeleton and was asked to walk at their preferred speed and follow the reference path as accurately as possible. Two 10-minute recordings were performed with the aid of visual feedback; one while using the HAPC and one while using the HPC with a 5-minute break between trials. Hybrid assistance was provided only for the muscles of the knee joint while for the hip joint only robotic assistance was provided  (surface electrodes were placed at the quadriceps and hamstrings as shown in Fig. \ref{fig:triadic_collaboration}b). The position of the ankles was adjusted according to the relative position in the gait cycle. The participant's ability to follow the reference path, the assistance they received and their estimated muscle fatigue were quantified and the sum of these parameters was calculated in order to evaluate the controller's ability to provide assistance as needed and prevent muscle fatigue. To facilitate the comparison between controllers, the results were normalised such that the costs for the HAPC are equal to 1.

\section{RESULTS}
\subsection{Simulation Results}
Fig. \ref{fig:sim_results} shows the controller's response to the simulated human behaviour described in section \ref{sec:sim_and_modelling}. It can be seen that for the first half of the simulation (indicated by a grey dotted line) the adaptive controller results in a reduction in exoskeleton stiffness, exoskeleton assistance (Fig. \ref{fig:sim_results}a), and FES stiffness through the adaptation of parameter $\gamma$ (Fig. \ref{fig:sim_results}b), while the trajectory error follows a periodic pattern (Fig. \ref{fig:sim_results}b) and muscle fatigue slowly increases (Fig. \ref{fig:sim_results}c). For the second half of the simulation the improved performance of the model reduces the tracking error (Fig. \ref{fig:sim_results}b) and the assistance received from both the exoskeleton (Fig. \ref{fig:sim_results}a) and the FES (Fig. \ref{fig:sim_results}b-c), which allows the muscles to recover (Fig. \ref{fig:sim_results}c). 

In Fig. \ref{fig:sim_results_adaptation_comparison} the performance of four different controllers is compared in response to the simulated
human behaviour described in section III-D. It can be seen that, compared to the HAPC, the EPC provides almost six times as much robotic assistance to keep the trajectory error low while the FPC results in higher trajectory error, higher stimulation intensity and more muscle fatigue than the HAPC. When comparing the HAPC to the HPC, a visible reduction in robotic assistance is observed while the rest of the parameters are affected less. This suggests that sufficient assistance is provided by the FES and that the controller's adaptation can reduce excessive assistive forces from the exoskeleton. These observations indicate that the hybridisation of robotic assistance and FES in this hierarchical order, and the controller's adaptation, provide a substantial reduction in the assistance provided to the user (at the expense of a higher tracking error when compared to the EPC), which overall reduces the sum of error, assistance and muscle fatigue, and implies an improved ability to provide assistance as needed.

\subsection{Experimental Results}
Fig. \ref{fig:HPC_ExpResults} shows the response of the HAPC to the recorded human behaviour and Fig. \ref{fig:S2_WeightedCostsResultsExperiment} provides a comparison between the HAPC and the HPC. Based on the results presented in Fig. \ref{fig:HPC_ExpResults}a it is evident that the controller's adaptation converged towards a lower exoskeleton stiffness than the baseline stiffness for both the hips and the knees\footnote{Results are shown for the left leg only as the results for the right leg follow similar patterns. These results can be provided upon request.}. This decrease in exoskeleton stiffness was also reflected in the assistance provided by the robot (Fig. \ref{fig:HPC_ExpResults}a). It is also evident that the exoskeleton stiffness of the knee joint converged to a smaller value than the exoskeleton stiffness of the hip joint (Fig. \ref{fig:HPC_ExpResults}a), which could be indicative of the cooperation between the robot and the FES. Similarly, a reduction in parameter $\gamma$ is observed, which reduces the stiffness of the FES controller (Fig. \ref{fig:HPC_ExpResults}b). This is also reflected in the stimulation provided by the adaptive controller to the knee (Fig. \ref{fig:HPC_ExpResults}c). As a result, a decreased muscle fatigue is observed (Fig. \ref{fig:S2_WeightedCostsResultsExperiment}), which for the left leg was more pronounced for the knee extensor muscles (Fig. \ref{fig:HPC_ExpResults}c). Meanwhile, the trajectory tracking error of the knee joint seems to vary only slightly (Fig. \ref{fig:HPC_ExpResults}b). 

It can be seen from Fig. \ref{fig:S2_WeightedCostsResultsExperiment}, that the sum of tracking error, assistance received and muscle fatigue, is smaller when the HAPC was used for both joints for both legs. For the hip joint, an average reduction of 49\% was recorded in the assistance provided by the exoskeleton across both legs, while the trajectory tracking error was higher by an average of 10\% across both legs when the HAPC was used. Similarly, for the knee joint, lower assistive forces were recorded from both the exoskeleton and the FES. The exoskeleton assistance was found to be an average of 58\% lower when the HAPC was used and the FES intensity was also lower by an average of 16\% across both legs. On the other hand, the tracking error was higher by 8\% for the left leg but lower by 8\% for the right leg when the HAPC was used. At the same time, the estimated fatigue of the knee muscles was reduced by an average of 37\% when the HAPC was used. This noticeable reduction in assistance and muscle fatigue, as compared to the less apparent change in tracking error, indicate the controller's ability to adapt to the user's performance in order to provide assistance as needed and preserve the fitness of their muscles.

\section{CONCLUSION}
In this manuscript a hybrid adaptive controller is presented that prioritises the voluntary human contributions and addresses muscle fatigue in a continuous and preventive fashion. This controller is designed to provide assistance as needed in order to encourage the active participation of the user, and preserve muscle fitness in order to prevent the premature termination of rehabilitation. It is shown that the controller is able to adapt to the user's movement, and adjust the levels of assistance provided by the robot and the FES in order to maintain an accurate tracking of the desired trajectory and preserve muscle fitness. These encouraging results indicate that the controller's hierarchical structure and adaptive components may be able to provide personalised assistance and prevent muscle fatigue which can be further investigated in the future, in a larger experimental study, including people with neurological disorders.   

\addtolength{\textheight}{-0cm}   % This command serves to balance the column lengths

%%%%%%%%%%%%%%%%%%%%%%%%%%%%%%%%%%%%%%%%%%%%%%%%%%%%%%%%%%%%%%%%%%%%%%%%%%%%%%%%

%%%%%%%%%%%%%%%%%%%%%%%%%%%%%%%%%%%%%%%%%%%%%%%%%%%%%%%%%%%%%%%%%%%%%%%%%%%%%%%%



% \section*{ACKNOWLEDGMENT}

%%%%%%%%%%%%%%%%%%%%%%%%%%%%%%%%%%%%%%%%%%%%%%%%%%%%%%%%%%%%%%%%%%%%%%%%%%%%%%%%

% \begin{thebibliography}{99}

% \end{thebibliography}
\bibliographystyle{IEEEtran}
% \bibliography{HILC_bibliography}

\input{root.bbl}

\end{document}

%% file: root.bbl

%% file: root.bbl
\begin{thebibliography}{10}
\providecommand{\url}[1]{#1}
\csname url@samestyle\endcsname
\providecommand{\newblock}{\relax}
\providecommand{\bibinfo}[2]{#2}
\providecommand{\BIBentrySTDinterwordspacing}{\spaceskip=0pt\relax}
\providecommand{\BIBentryALTinterwordstretchfactor}{4}
\providecommand{\BIBentryALTinterwordspacing}{\spaceskip=\fontdimen2\font plus
\BIBentryALTinterwordstretchfactor\fontdimen3\font minus \fontdimen4\font\relax}
\providecommand{\BIBforeignlanguage}[2]{{%
\expandafter\ifx\csname l@#1\endcsname\relax
\typeout{** WARNING: IEEEtran.bst: No hyphenation pattern has been}%
\typeout{** loaded for the language `#1'. Using the pattern for}%
\typeout{** the default language instead.}%
\else
\language=\csname l@#1\endcsname
\fi
#2}}
\providecommand{\BIBdecl}{\relax}
\BIBdecl

\bibitem{Campagnini2022}
S.~Campagnini, P.~Liuzzi, A.~Mannini, R.~Riener, and M.~C. Carrozza, ``{Effects of control strategies on gait in robot-assisted post-stroke lower limb rehabilitation: a systematic review},'' \emph{Journal of NeuroEngineering and Rehabilitation}, vol.~19, no.~1, pp. 1--16, 2022.

\bibitem{Shi2019a}
B.~Shi, X.~Chen, Z.~Yue, S.~Yin, Q.~Weng, X.~Zhang, J.~Wang, and W.~Wen, ``{Wearable Ankle Robots in Post-stroke Rehabilitation of Gait: A Systematic Review},'' \emph{Frontiers in Neurorobotics}, vol.~13, no. August, pp. 1--16, 2019.

\bibitem{Fang2020}
C.-Y. Fang, J.-L. Tsai, G.-S. Li, A.~S.-Y. Lien, and Y.-J. Chang, ``{Effects of Robot-Assisted Gait Training in Individuals with Spinal Cord Injury: A Meta-analysis},'' \emph{BioMed Research International}, vol. 2020, pp. 1--13, mar 2020.

\bibitem{Hayes2018}
S.~C. Hayes, C.~R. {James Wilcox}, H.~S. {Forbes White}, and N.~Vanicek, ``{The effects of robot assisted gait training on temporal-spatial characteristics of people with spinal cord injuries: A systematic review},'' \emph{Journal of Spinal Cord Medicine}, vol.~41, no.~5, pp. 529--543, 2018.

\bibitem{Tedla2019}
J.~S. Tedla, S.~Dixit, K.~Gular, and M.~Abohashrh, ``{Robotic-Assisted Gait Training Effect on Function and Gait Speed in Subacute and Chronic Stroke Population: A Systematic Review and Meta-Analysis of Randomized Controlled Trials},'' \emph{European Neurology}, vol.~81, no. 3-4, pp. 103--111, 2019.

\bibitem{Barbuto2019a}
S.~Barbuto and J.~Stein, \emph{{Rehabilitation Robotics for Stroke}}.\hskip 1em plus 0.5em minus 0.4em\relax Elsevier Inc., 2019.

\bibitem{Swinnen2014}
E.~Swinnen, D.~Beckw{\'{e}}e, R.~Meeusen, J.~P. Baeyens, and E.~Kerckhofs, ``{Does robot-assisted gait rehabilitation improve balance in stroke patients? a systematic review},'' \emph{Topics in Stroke Rehabilitation}, vol.~21, no.~2, pp. 87--100, 2014.

\bibitem{Anaya2018}
F.~Anaya, P.~Thangavel, and H.~Yu, ``{Hybrid FES–robotic gait rehabilitation technologies: a review on mechanical design, actuation, and control strategies},'' \emph{International Journal of Intelligent Robotics and Applications}, vol.~2, no.~1, pp. 1--28, 2018.

\bibitem{Del-Ama2012}
A.~J. Del-Ama, A.~D. Koutsou, J.~C. Moreno, A.~De-los Reyes, N.~Gil-Agudo, and J.~L. Pons, ``{Review of hybrid exoskeletons to restore gait following spinal cord injury},'' \emph{The Journal of Rehabilitation Research and Development}, vol.~49, no.~4, p. 497, 2012.

\bibitem{Peckham2005}
P.~H. Peckham and J.~S. Knutson, ``{Functional Electrical Stimulation for Neuromuscular Applications},'' \emph{Annual Review of Biomedical Engineering}, vol.~7, no.~1, pp. 327--360, aug 2005.

\bibitem{Marquez-Chin2020}
C.~Marquez-Chin and M.~R. Popovic, ``{Functional electrical stimulation therapy for restoration of motor function after spinal cord injury and stroke: A review},'' \emph{BioMedical Engineering Online}, vol.~19, no.~1, pp. 1--25, 2020.

\bibitem{PopovicM2016}
M.~R. Popovic, K.~Masani, and S.~Micera, ``{Functional Electrical Stimulation Therapy: Recovery of Function Following Spinal Cord Injury and Stroke},'' in \emph{Neurorehabilitation Technology}.\hskip 1em plus 0.5em minus 0.4em\relax Cham: Springer International Publishing, 2016, vol.~16, pp. 513--532.

\bibitem{Ibitoye2016b}
M.~O. Ibitoye, N.~A. Hamzaid, N.~Hasnan, A.~K.~A. Wahab, and G.~M. Davis, ``{Strategies for rapid muscle fatigue reduction during FES exercise in individuals with spinal cord injury: A systematic review},'' \emph{PLoS ONE}, vol.~11, no.~2, pp. 1--28, 2016.

\bibitem{Bickel2011}
C.~S. Bickel, C.~M. Gregory, and J.~C. Dean, ``{Motor unit recruitment during neuromuscular electrical stimulation: a critical appraisal},'' \emph{European Journal of Applied Physiology}, vol. 111, no.~10, pp. 2399--2407, oct 2011.

\bibitem{Lynch2008}
C.~L. Lynch and M.~R. Popovic, ``{Functional Electrical Stimulation},'' \emph{IEEE Control Systems}, vol.~28, no.~2, pp. 40--50, 2008.

\bibitem{Schauer2017}
T.~Schauer, ``{Sensing motion and muscle activity for feedback control of functional electrical stimulation: Ten years of experience in Berlin},'' \emph{Annual Reviews in Control}, vol.~44, pp. 355--374, 2017.

\bibitem{Goldfarb1996}
M.~Goldfarb and W.~K. Durfee, ``{Design of a controlled-brake orthosis for FES-aided gait},'' \emph{IEEE Transactionl on Rehabilitation Engineering}, vol.~4, no.~1, pp. 13--24, 1996.

\bibitem{Kobetic2009}
R.~Kobetic, C.~S. To, J.~R. Schnellenberger, M.~L. Audu, T.~C. Bulea, R.~Gaudio, G.~Pinault, S.~Tashman, and R.~J. Triolo, ``{Development of hybrid orthosis for standing, walking, and stair climbing after spinal cord injury},'' \emph{The Journal of Rehabilitation Research and Development}, vol.~46, no.~3, p. 447, 2009.

\bibitem{Sharma2014}
N.~Sharma, V.~Mushahwar, and R.~Stein, ``{Dynamic optimization of FES and orthosis-based walking using simple models},'' \emph{IEEE Transactions on Neural Systems and Rehabilitation Engineering}, vol.~22, no.~1, pp. 114--126, 2014.

\bibitem{Stauffer2009b}
Y.~Stauffer, Y.~Allemand, M.~Bouri, J.~Fournier, R.~Clavel, P.~Metrailler, R.~Brodard, and F.~Reynard, ``{The WalkTrainer - A new generation of walking reeducation device combining orthoses and muscle stimulation},'' \emph{IEEE Transactions on Neural Systems and Rehabilitation Engineering}, vol.~17, no.~1, pp. 38--45, 2009.

\bibitem{Ha2016d}
K.~H. Ha, S.~A. Murray, and M.~Goldfarb, ``{An Approach for the Cooperative Control of FES With a Powered Exoskeleton During Level Walking for Persons With Paraplegia},'' \emph{IEEE Transactions on Neural Systems and Rehabilitation Engineering}, vol.~24, no.~4, pp. 455--466, 2016.

\bibitem{Del-Ama2014a}
A.~J. Del-Ama, {\'{A}}.~Gil-Agudo, J.~L. Pons, and J.~C. Moreno, ``{Hybrid FES-robot cooperative control of ambulatory gait rehabilitation exoskeleton},'' \emph{Journal of NeuroEngineering and Rehabilitation}, vol.~11, no.~1, pp. 1--15, 2014.

\bibitem{Kirsch2017}
N.~Kirsch, N.~Alibeji, and N.~Sharma, ``{Nonlinear model predictive control of functional electrical stimulation},'' \emph{Control Engineering Practice}, vol.~58, pp. 319--331, 2017.

\bibitem{Molazadeh2021}
V.~Molazadeh, Q.~Zhang, X.~Bao, B.~E. Dicianno, and N.~Sharma, ``{Shared Control of a Powered Exoskeleton and Functional Electrical Stimulation Using Iterative Learning},'' \emph{Frontiers in Robotics and AI}, vol.~8, no. November, pp. 1--13, 2021.

\bibitem{Alibeji2018a}
N.~A. Alibeji, V.~Molazadeh, B.~E. Dicianno, and N.~Sharma, ``{A control scheme that uses dynamic postural synergies to coordinate a hybrid walking neuroprosthesis: Theory and experiments},'' \emph{Frontiers in Neuroscience}, vol.~12, no. APR, pp. 1--15, 2018.

\bibitem{Vallery2005}
H.~Vallery, T.~St{\"{u}}tzle, M.~Buss, and D.~Abel, ``{Control of a hybrid motor prosthesis for the knee joint},'' \emph{IFAC Proceedings Volumes (IFAC-PapersOnline)}, vol.~38, no.~1, pp. 76--81, 2005.

\bibitem{Romero-Sanchez2019a}
F.~Romero-S{\'{a}}nchez, J.~Bermejo-Garc{\'{i}}a, J.~Barrios-Muriel, and F.~J. Alonso, ``{Design of the cooperative actuation in hybrid orthoses: A theoretical approach based on muscle models},'' \emph{Frontiers in Neurorobotics}, vol.~13, no. July, pp. 1--15, 2019.

\bibitem{Blank2014}
A.~A. Blank, J.~A. French, A.~U. Pehlivan, and M.~K. O'Malley, ``{Current Trends in Robot-Assisted Upper-Limb Stroke Rehabilitation: Promoting Patient Engagement in Therapy},'' \emph{Current Physical Medicine and Rehabilitation Reports}, vol.~2, no.~3, pp. 184--195, 2014.

\bibitem{Paolucci2012}
S.~Paolucci, A.~{Di Vita}, R.~Massicci, M.~Traballesi, I.~Bureca, A.~Matano, M.~Iosa, and C.~Guariglia, ``{Impact of participation on rehabilitation results: a multivariate study.}'' \emph{European journal of physical and rehabilitation medicine}, vol.~48, no.~3, pp. 455--66, sep 2012.

\bibitem{Kahn2006}
L.~E. Kahn, P.~S. Lum, W.~Z. Rymer, and D.~J. Reinkensmeyer, ``{Robot-assisted movement training for the stroke-impaired arm: Does it matter what the robot does?}'' \emph{Journal of Rehabilitation Research and Development}, vol.~43, no.~5, pp. 619--629, 2006.

\bibitem{Kaelin-Lane2005}
A.~Kaelin-Lane, L.~Sawaki, and L.~G. Cohen, ``{Role of voluntary drive in encoding an elementary motor memory},'' \emph{Journal of Neurophysiology}, vol.~93, no.~2, pp. 1099--1103, 2005.

\bibitem{Gordon2023}
D.~F.~N. Gordon, A.~Christou, T.~Stouraitis, M.~Gienger, and S.~Vijayakumar, ``{Adaptive assistive robotics: a framework for triadic collaboration between humans and robots},'' \emph{Royal Society Open Science}, vol.~10, no.~6, jun 2023.

\bibitem{Christou2022}
A.~Christou, D.~Gordon, T.~Stouraitis, and S.~Vijayakumar, ``{Designing Personalised Rehabilitation Controllers using Offline Model-Based Optimisation},'' in \emph{2022 IEEE International Conference on Robotics and Biomimetics (ROBIO)}.\hskip 1em plus 0.5em minus 0.4em\relax IEEE, dec 2022, pp. 148--155.

\bibitem{Duschau-Wicke2010b}
A.~Duschau-Wicke, J.~{Von Zitzewitz}, A.~Caprez, L.~L{\"{u}}nenburger, and R.~Riener, ``{Path control: A method for patient-cooperative robot-aided gait rehabilitation},'' \emph{IEEE Transactions on Neural Systems and Rehabilitation Engineering}, vol.~18, no.~1, pp. 38--48, 2010.

\bibitem{Riener1998}
R.~Riener and T.~Fuhr, ``{Patient-driven control of FES-supported standing up: A simulation study},'' \emph{IEEE Transactions on Rehabilitation Engineering}, vol.~6, no.~2, pp. 113--124, 1998.

\bibitem{GFOHLER2004}
M.~Gfohler, T.~Angeli, and P.~Lugner, ``{Modeling of artificially activated muscle and application to FES cycling},'' \emph{Journal of Mechanics in Medicine and Biology}, vol.~04, no.~01, pp. 77--92, 2004.

\bibitem{Riener1996}
R.~Riener, J.~Quintern, and G.~Schmidt, ``{Biomechanical model of the human knee evaluated by neuromuscular stimulation},'' \emph{Journal of Biomechanics}, vol.~29, no.~9, pp. 1157--1167, 1996.

\bibitem{Sharma2017b}
N.~Sharma, N.~A. Kirsch, N.~A. Alibeji, and W.~E. Dixon, ``{A non-linear control method to compensate for muscle fatigue during neuromuscular electrical stimulation},'' \emph{Frontiers Robotics AI}, vol.~4, no. DEC, 2017.

\bibitem{Seth2018}
A.~Seth, J.~L. Hicks, T.~K. Uchida, A.~Habib, C.~L. Dembia, J.~J. Dunne, C.~F. Ong, M.~S. DeMers, A.~Rajagopal, M.~Millard, S.~R. Hamner, E.~M. Arnold, J.~R. Yong, S.~K. Lakshmikanth, M.~A. Sherman, J.~P. Ku, and S.~L. Delp, ``{OpenSim: Simulating musculoskeletal dynamics and neuromuscular control to study human and animal movement},'' \emph{PLOS Computational Biology}, vol.~14, no.~7, p. e1006223, 2018.

\bibitem{Millard2013}
M.~Millard, T.~Uchida, A.~Seth, and S.~L. Delp, ``{Flexing computational muscle: Modeling and simulation of musculotendon dynamics},'' \emph{Journal of Biomechanical Engineering}, vol. 135, no.~2, pp. 1--11, 2013.

\end{thebibliography}
